\documentclass[10pt,letterpaper]{article}

\usepackage{cogsci}

\cogscifinalcopy 

\usepackage{xcolor}

\usepackage{pslatex}
\usepackage{apacite}
\usepackage{float} 

\usepackage{graphicx}
\usepackage{amsmath, amssymb, mathtools, amsfonts, amsthm}

\title{Can Humans Do Less-Than-One-Shot Learning?}
 
\author{{\large \bf Maya Malaviya\textsuperscript{1,*}, Ilia Sucholutsky\textsuperscript{2,*},  Kerem Oktar\textsuperscript{1}, Thomas L. Griffiths\textsuperscript{1,2}} \\
  \textsuperscript{1}Department of Psychology, Princeton University\\
  \textsuperscript{2}Department of Computer Science, Princeton University\\
  $\texttt{\{mayamala, is2961, oktar, tomg\}@princeton.edu}$ \\
  \textsuperscript{*}Equal contribution.} 

\begin{document}

\maketitle

\begin{abstract}
Being able to learn from small amounts of data is a key characteristic of human intelligence, but exactly {\em how} small? 
In this paper, we introduce a novel experimental paradigm that allows us to examine classification in an extremely data-scarce setting, asking whether humans can learn more categories than they have exemplars (i.e., can humans do ``less-than-one shot'' learning?). An experiment conducted using this paradigm reveals that people are capable of learning in such settings, and provides several insights into underlying mechanisms. First, people can accurately infer and represent high-dimensional feature spaces from very little data. Second, having inferred the relevant spaces, people use a form of prototype-based categorization (as opposed to exemplar-based) to make categorical inferences. Finally, systematic, machine-learnable patterns in responses indicate that people may have efficient inductive biases for dealing with this class of data-scarce problems. 

\end{abstract}

\section{Introduction}
Machine learning (ML) systems now approach or exceed human performance across a wide variety of tasks \cite{lecunDeepLearning2015}. Much of this progress is attributable to the use of highly flexible algorithms in very data-rich settings---most benchmark tasks in classification, for instance, provide hundreds or thousands of examples per class \cite<e.g.>{krizhevsky2012imagenet}. In contrast, human intelligence is characterized by its impressive data efficiency. Both children and adults can effectively learn complex concepts in few- or ``one-shot'' settings (i.e., after seeing a few or one examples; see below for a detailed review). This contrast in the efficiency of learning has inspired recent ML research in few-shot learning \cite<for a review, see>{wangGeneralizingFewExamples2020}. A recent finding suggests that it is technically possible to learn in less-than-one-shot (LO-shot) settings \cite<e.g., learning \textit{N} classes from $M<N$ samples, see>{sucholutskyLessOneShot2020}. It is currently unknown whether humans can learn in such a setting. Moreover, no existing experimental paradigms are fit to investigate this question, as LO-shot learning requires using  ``soft labels'' (which provide graded category membership) as opposed to the ``hard'' discrete assignments that are typically used.

In this paper, we introduce a novel experimental paradigm that allows us to test LO-shot learning in humans, and present the results of an experiment conducted in this paradigm. These results are both the first demonstration of successful LO-shot learning in humans, and provide the following novel insights into the psychological mechanisms enabling this capacity. First, the distribution of participants' responses indicates that they accurately infer the structure and statistics of the data-generating process, without any explicit instruction. Second, comparing participant data with the predictions of computational models suggests that people use a form of prototype-based categorization (as opposed to exemplar-based). Third, participant responses show systematic, machine-learnable patterns, which suggests that people may have efficient inductive biases for dealing with this class of data-scarce problems. 
Combined, these results demonstrate the power of our paradigm to reveal novel insights into cognition and shed light on existing debates in psychology. 

\section{Few-Shot Learning in Humans}%
A fundamental question in cognitive science is how people come to know so much about the world from such little input \cite{russell1948human}. There is much evidence to suggest that people can learn from few examples: toddlers learn and generalize the functions of artifacts \cite{caslerYoungChildrenRapid2005} and meanings of words  \cite{careyAcquiringSingleNew1978,careyFastMapping2010,bloomHowChildrenLearn2000} after exposure to just one instance, whereas adults can learn novel recursive  \cite{lake2020people}, compositional \cite{lake2019human}, lexical \cite{coutancheFastMappingRapidly2014}, and visual \cite{lakeOneShotLearning2011} concepts in similar settings. But what underlies such impressive efficiency?

Early theorists proposed that human knowledge is too complex to be learned from limited input, and thus must be innate \cite<e.g.,>{chomsky1986knowledge}. But mounting evidence suggests that statistical inference can be sufficient \cite{xuInfantsAreRational2013,tenenbaumHowGrowMind2011a} when supported by inductive biases resulting from hierarchical inference \cite{yuanLearningGenerativePrinciples2020}, active allocation of attention \cite{smithObjectNameLearning2002}, and well-calibrated priors \cite{ruleLeveragingPriorConcept2021}. LO-shot learning is a problem that contributes to this debate by highlighting the surprising complexity of statistical inferences that can be drawn from rich and scarce data. It might seem impossible to learn a category without seeing any examples from it---yet such inferences are quite common. 

\section{Computational Principles of LO-shot Learning}
In the machine-learning literature, few-shot learning methods can be seen as a response to the growing problem of deep learning requiring massive models to be trained on massive datasets. This scaling trend has led to novel deep learning technologies being inaccessible to the broader research community due to the data requirements and computational cost, as well as potentially being harmful to the environment due to their energy cost~\cite{energy}. While few-shot learning can greatly reduce the required number of training examples over typical deep learning techniques, this number still scales at least linearly with the number of classes present in the dataset. For example, one-shot learning still requires one training example for every class in the dataset~\cite{fei2006one}. 

\begin{figure}[b!]
\begin{align*}
\begin{bmatrix}
           0.6 \\
           0.3 \\
           0.1 \\
\end{bmatrix}
\xrightarrow[\text{and rest to 0}]{\text{Set largest to 1}}
\begin{bmatrix}
           1 \\
           0 \\
           0 \\
\end{bmatrix}
\end{align*}
\caption{Soft and hard labels. The vector on the left shows a soft label---a probability distribution over a stimuli's membership to each class. In this example, the soft label suggests 60\%, 30\%, and 10\% probability of membership to the first, second, and third classes, respectively. The vector on the right shows a hard label, which indicates a stimulus belongs to a single class. In this example, the hard label suggests strict membership in the first class. Hard labels can be derived from soft labels by using an argmax function (i.e. setting the highest probability element to 1, and the rest to 0).}
\label{soft_label_example}
\end{figure}

LO-shot learning emerged as an attempt to probe the limits of few-shot learning and determine whether it was possible to have sub-linear scaling of the required number of training examples. The first evidence of LO-shot learning in neural networks came about as a by-product of research into ``dataset distillation''---a method for ``distilling'' large training datasets into small synthetic datasets that will still train models to comparable test accuracies~\cite{DD}. \citeA{SLDD} showed that the size of the synthetic ``distilled'' dataset could be reduced to below one training example per class if each synthetic training example were assigned a learnable soft label instead of the usual hard labels (see Figure~\ref{soft_label_example} for an explanation of hard and soft labels). They demonstrated this by showing a neural network can accurately recognize handwritten images of the ten digits (0-9) after being trained on a total of just five synthetic images. Further research into LO-shot learning focused on using analytical methods to test the theoretical limits of learning with small data and showed that the number of soft-labeled training examples required to learn $K$ classes was actually a function of the geometry of the data rather than of $K$~\cite{sucholutskyLessOneShot2020}. A key result from this analysis was that just two soft-label training examples are sufficient to characterize any finite number of classes, if those classes are all approximately situated on the same one-dimensional manifold.  \citeA{Sucholutsky2021OneLT} propose algorithmic methods for detecting and exploiting such manifolds in datasets. 

In this paper, we propose a framework for testing whether humans also have strategies for disentangling class information in small data regimes. In particular, in this initial study we test the simplest formulation of the result above: can humans learn to characterize three classes lying on a 1-D manifold when given just two soft-label training examples? 

\section{Methods}

\subsection{Participants} 
Participants were 70 adults (25 male, 43 female, 2 non-binary, mean age = 36) recruited from the Prolific platform in exchange for monetary compensation (\$2.00 for a 15-minute experiment). Participants were to be excluded if they gave the same response for each trial throughout the experiment, which could possibly indicate lapsed attention. However, no participants were eliminated from the original sample as none of them met this pre-determined exclusion criterion.

\subsection{Stimuli}

Testing LO-Shot learning in humans can be difficult, because it requires eliciting people's notions of a category that they have never seen before. Our paradigm proposes that we can glean this information by asking participants to classify a stimulus into one of many unseen categories.

The stimuli for this experiment were adapted from \citeA{sanborn2008MCMCwithpeople}. These stick-figure models of quadrupeds have 9 distinct, continuous features, so they are characterized by a nine-dimensional space. The stimuli were generated by manually instantiating feature values for two stick figures, then taking linear combinations of these values to produce more stick figures with new feature values. Because we only took linear combinations of the feature values, which were scaled by their respective possible ranges, the stimuli all lie on a 1-D manifold in a 9-D feature space.

\begin{figure}[t!]
    \begin{center}
    \includegraphics[width=\columnwidth]{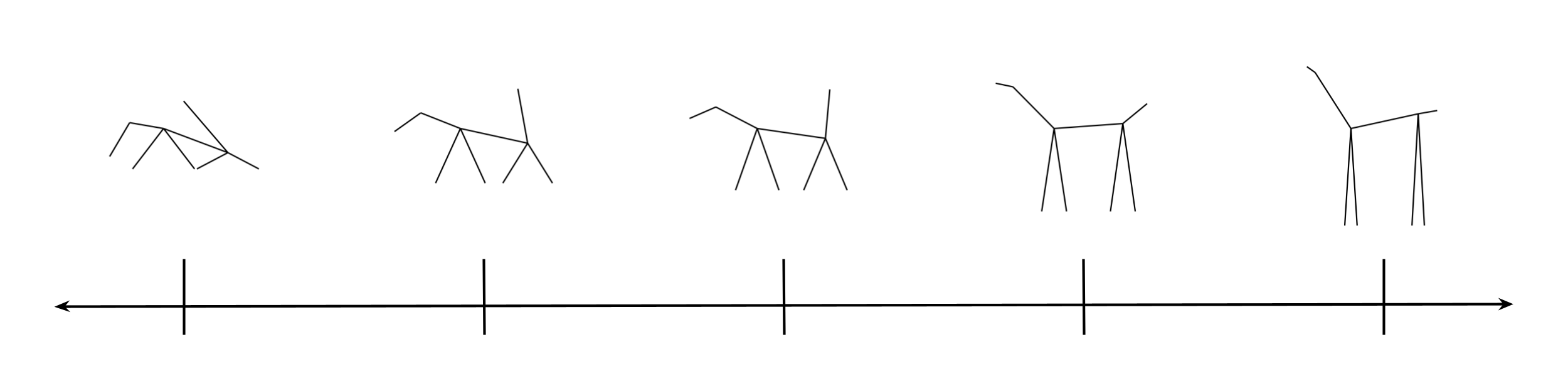}
    \end{center}
\caption{Generated stick figures along a 1-D manifold.} 
\label{fig:stickfigstim}
\end{figure}

We also repeated the stimulus generation procedure, so that we could determine if results were robust across two different manifolds (Manifold 1 and Manifold 2). Each manifold contains 20 generated stimuli. To see a subset of generated stick figures and examples of how the features can change across the manifold, see Figure~\ref{fig:stickfigstim}.

To make the these stimuli intuitive to understand, we framed the stick figures as models that paleontologists use to summarize dinosaur fossil structure. Therefore, we referred to the stick figures as dinosaurs throughout the experiment.

The task we built took inspiration from LO-shot learning methods in ML by adapting the concept of soft labels, which are a way to represent an object's simultaneous membership to several classes to a human setting (see Figure~\ref{soft_label_example} for an explanation of soft labels). In order to make this concept intuitive to participants, we gave them soft labels in the form of genetic information. Specifically, we indicated that dinosaurs had a certain amount of genetic overlap with three unseen species: Species 1, Species 2, and Species 3. If subjects build characterizations for Species 1, 2, and 3 when given only two soft-labelled dinosaurs, we can successfully probe whether humans are capable of LO-shot learning in this paradigm.

We built soft-label data in the form of soft-label pairs (SLPs) because we planned to show two dinosaurs, each with their own genetic information pertaining to Species 1, 2, and 3. SLPs were chosen according to the following criteria: a) they should use probabilities that are easily interpretable in the context of genetic background (i.e. 0, 0.25, 0.5, 0.75, 1); b) they should be informative about all three classes; c) they should form valid probability distributions; and d) they should capture a diverse set of possible configurations.
The final set of SLPs used in this study are listed in Table~\ref{slp-table}.

\begin{table}[t!]
\begin{center} 
\caption{Soft Label Pairs (SLPs)} 
\label{slp-table} 
\vskip 0.12in
\begin{tabular}{c c c} 
\hline
    & Dinosaur 1   &  Dinosaur 2 \\
\hline
1 &  [0\%, 0\%, 100\%] & [25\%, 50\%, 25\%]\\
2 &  [0\%, 0\%, 100\%] & [25\%, 75\%, 0\%]\\
4 &  [0\%, 25\%, 75\%] & [25\%, 0\%, 75\%]\\
5 &  [0\%, 25\%, 75\%] & [25\%, 25\%, 50\%]\\
6 &  [0\%, 25\%, 75\%] & [25\%, 50\%, 25\%]\\
7 &  [0\%, 25\%, 75\%] & [25\%, 75\%, 0\%]\\
8 &  [0\%, 25\%, 75\%] & [50\%, 0\%, 50\%]\\
9 &  [0\%, 25\%, 75\%] & [50\%, 25\%, 25\%]\\
10 &  [0\%, 25\%, 75\%] & [50\%, 50\%, 0\%]\\
11 &  [0\%, 25\%, 75\%] & [75\%, 25\%, 0\%]\\
12 & [0\%, 50\%, 50\%] & [25\%, 25\%, 50\%]\\
13 &  [0\%, 50\%, 50\%] & [50\%, 0\%, 50\%]\\
14 &  [0\%, 50\%, 50\%] & [50\%, 25\%, 25\%]\\
16 &  [25\%, 25\%, 50\%] & [25\%, 50\%, 25\%]\\
\hline
\end{tabular} 
\end{center} 
\end{table}

\subsection{Experimental Procedure}
Participants were randomly assigned into one of 14 conditions. Each group was introduced to a different set of soft labels which were assigned to Dinosaurs 1 and 2 for the whole experiment. All 14 SLPs are listed in Table~\ref{slp-table}.

To introduce the experiment, we presented participants with a vignette, framing the stick-figure stimuli as models that paleontologists use to compare the anatomy of dinosaur fossils they have uncovered at dig sites. The soft labels were explained as results of a long and expensive DNA analysis which revealed that these labeled dinosaurs are the descendants of three previously unknown species of dinosaur. Therefore, we can offer a rationale for the classification task: ``the scientists believe that by comparing the stick-figure model of a new fossil against those of the first two fossils and their genetic information, it may be possible to deduce the closest relative of the new fossil.'' The participants were then presented with two of the dinosaur fossil models (labeled Dinosaur 1 and Dinosaur 2) and genetic information for each. They then saw a third dinosaur (Dinosaur 3) and were asked to use the appearance of all of the dinosaurs and the genetic information from Dinosaurs 1 and 2 to conclude which species Dinosaur 3 is most closely related to.

After the participant submitted their response, the trial was repeated, but Dinosaur 3 was a new, unseen stick figure from the manifold. Thus, participants completed 20 trials per manifold. Because we generated two manifolds, mid-way through the experiment participants were alerted that they would be changing to new dig sites with new dinosaurs. They proceeded to repeat the same process but with dinosaurs generated from the second manifold. The order of Manifolds 1 and 2 was randomized for each participant. Note that while Dinosaurs 1 and 2 change across the manifolds, the genetic information presented remains the same for each participant.
\begin{figure}[t!]
    \begin{center}
    \includegraphics[width=\columnwidth]{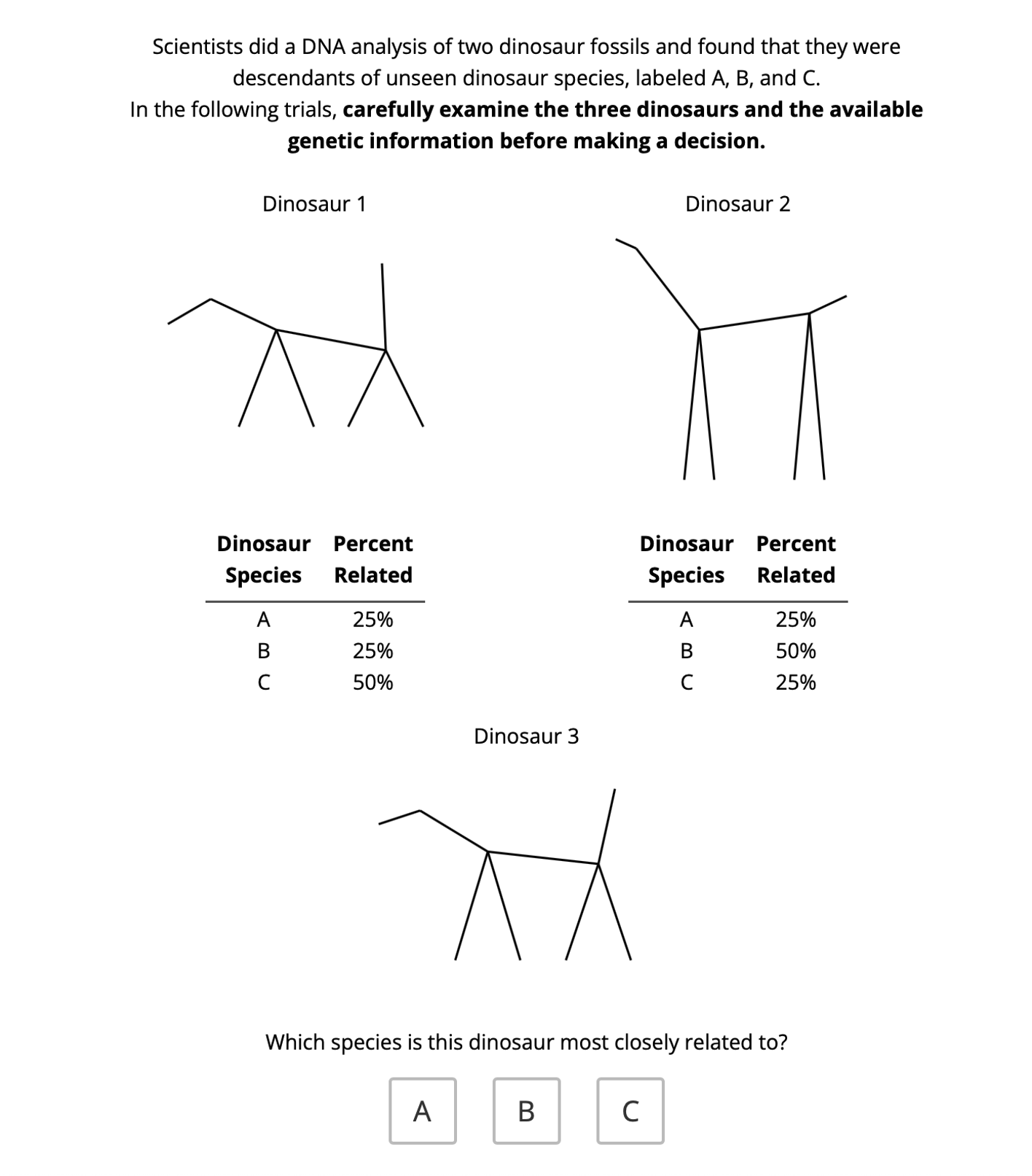}
    \end{center}
\caption{A screen capture from the experiment. In this trial, participants classified a dinosaur from Manifold 1, with genetic information from SLP 16.} 
\label{fig:testtrial}
\end{figure}
\begin{figure*}[t!]
    \begin{center}
    \includegraphics[width=\textwidth]{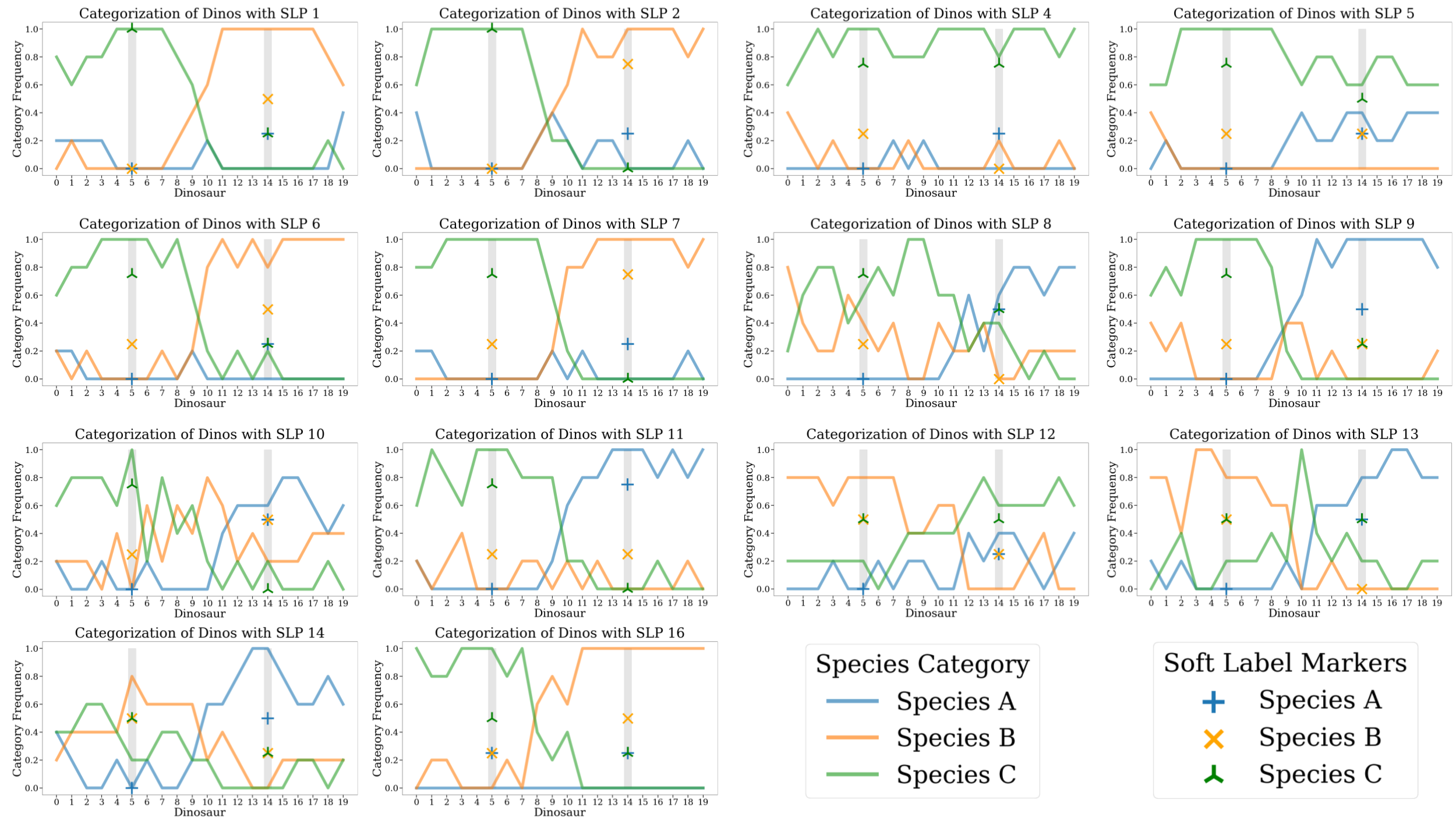}
    \end{center}
\caption{Distributions of participant responses for every SLP. Each color represents one of the possible three classification responses. The grey columns represent the locations along the manifold of Dinosaur 1 and Dinosaur 2 shown to participants. Along the grey columns, colored markers represent the genetic information associated with the dinosaur at that location.} 
\label{fig:subjcateg}
\end{figure*}
\section{Results}
Our analysis of our data had three aims.
First, to determine that a) participants are actually performing some sort of meaningful learning from the soft labels in our experiment and that b) strategies are robust across different participants. 
Second, to explore which approach to classification participants may be using when developing their strategies in this experiment. 
Finally, to show that machine-learning models can predict human behavior in this experiment. 
\subsection{Learning from Soft Labels in LO-Shot Settings}
One of the benefits of our proposed framework is that it enables evaluation of the human ability to disentangle class information in LO-shot regimes. We performed several analyses which suggest that participants are discovering meaningful, reproducible strategies in this setting. 

First, we can qualitatively observe in the population distributions in Figure~\ref{fig:subjcateg} that participants have non-trivial posteriors over all three classes over the entire manifold and that these posteriors vary with changes in the presented SLP. 
When analyzed at a subject-level, we find that the conclusion is the same: individual participants have non-trivial posteriors over all three classes that vary with position of the target dinosaur along the manifold. 
Furthermore, we tested the variance within every SLP (i.e. a 20-by-3 contingency table) using Pearson's chi-squared test and find that, even after applying the Bonferroni correction for multiple comparisons, the results are statistically significant for all ($\chi^2(38)\geq90$, $p<.001$) but two SLPs (SLP-4: $\chi^2(38)=41.7$, $p=0.312$; SLP-5: $\chi^2(38)=47.0$, $p=0.149$). We also test the effect of choice of SLP (i.e. a 14-by-3 table as we aggregate over the manifold for each SLP) and find that the results are also statistically significant ($\chi^2(26)=721.9$, $p<.001$). 

As mentioned above, each participant completed 40 trials on their assigned SLP, 20 of which are with Manifold 1 and 20 with Manifold 2. Since the soft labels remain constant between these two sets of 20 trials (only the presented pair of dinosaurs changes), we can compute within-subject agreement (WSA) by comparing a participant's responses on the Manifold 1 to their responses on Manifold 2. Similarly, we can calculate between-subject agreement (BSA) by comparing responses of all pairs of participants assigned to the same SLP. Across our dataset, we find that WSA is fairly high with an average of 74.07\% while a random selection strategy has an expected WSA of 33.33\% (binomial test, 
$p < .001$). This suggests that participants develop a consistent strategy and that the induced posterior is based on a relative comparison of the stimuli (i.e. the manifold described by the SLP) rather than the absolute stimuli. We find that BSA is 68.18\% (chance value is still 33.33\%; binomial test, 
$p < 0.00001$) which suggests that different participants may be discovering the same strategies for disentangling classes.

\begin{figure}[b!]
    \centering
    \includegraphics[width=0.49\columnwidth]{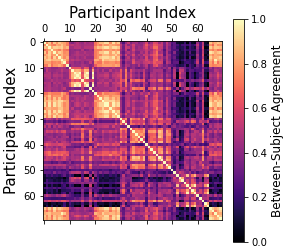}
    \includegraphics[width=0.49\columnwidth]{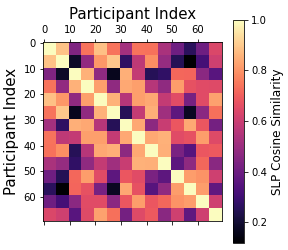}
    \caption{Comparing the similarity matrix of participant responses to the similarity matrix of the SLPs they were shown. The left shows a heatmap of pairwise between-subject agreement (i.e. the similarity between participant responses). The right shows a heatmap of cosine similarity between the actual SLPs assigned to these participants.}
    \label{fig:heatmaps}
\end{figure}

In order to further test whether participants' strategies make use of the soft-label information, we compare whether the BSA between pairs of participants is correlated with the similarity between the two SLPs those participants were shown. To compute similarity between pairs of SLPs, we flatten each SLP into a single vector of length six (three probability values from each of the two soft labels in an SLP) and compute cosine similarity (dot product divided by the product of the norms) between every pair of such vectors.  

In Figure~\ref{fig:heatmaps}, we present a heatmap of pairwise BSA across all participants alongside a heatmap of cubed pairwise cosine similarities of the SLPs presented to each participant (cubing the similarities helps better distinguish the structure). The Pearson correlation coefficient between the upper-triangular entries of the two matrices is $r(2483)=0.566$, $p<0.001$ which suggests moderate linear correlation. Based on the well-defined structures visible in the BSA heatmap, we conclude that subjects' strategies make significant use of the provided soft labels but that a similarity metric other than cosine similarity may be more predictive of BSA.

\subsection{Prototype and Exemplar Models}
A comprehensive overview of psychological models of categorization is outside the scope of this paper \cite<instead, see>{murphyBigBookConcepts2002,kruschkeModelsCategorization2008}. However, we will introduce the two main classes of models that have been used in the literature and analyze which better matches human behavior in LO-shot settings. Exemplar models \cite<e.g.,>{medin1978context,nosofsky1987attention} typically store each instance of a category that has been encountered thus far, and categorize new stimuli according to the similarity of the stimulus to every exemplar in memory. Prototype models \cite<e.g.,>{reed1972pattern}
operate analogously, but derive summary representations of learned categories and compute similarity with these ``prototypes'' (instead of computing it across all instances). There are few restrictions on what constitutes a summary representation---it could be the modal exemplar for a category, the central exemplar, or an ``ideal'' \cite<see>{murphyBigBookConcepts2002}.

We compare our study population results against three parameter-free classification models: a) a prototype model that first fits three hard-label prototypes (one for each class) to the data and then uses them with a distance-weighted 3-nearest neighbor classifier for prediction; b) a 1-nearest neighbor (1NN) exemplar model that copies the nearest labeled dinosaur when predicting probabilities at a target point; and c) a distance-weighted 2-nearest neighbor (2NN) exemplar model. When making a prediction, the distance-weighted $k$-nearest neighbor models that we used output the sum of the $k$ provided labels (hard or soft) weighted by their inverse square distance from the target point.
\begin{figure}[t!]
    \centering
    \includegraphics[width=\columnwidth]{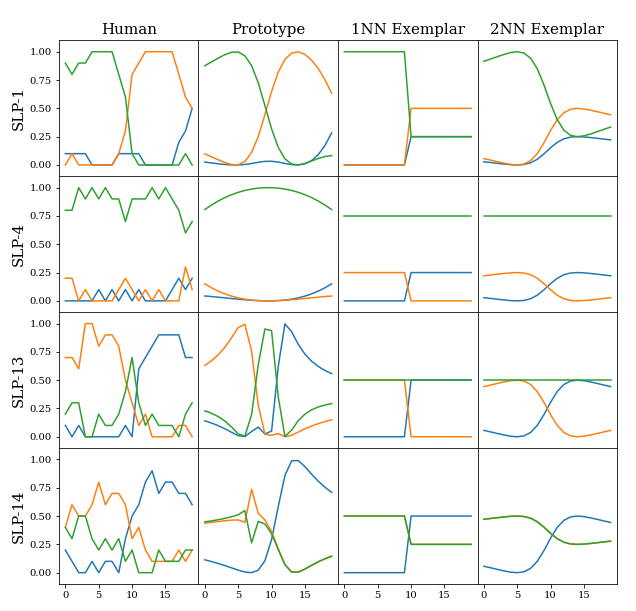}
    \caption{Comparing classification strategies to population results for four SLPs. The first column shows the distribution in our population of human participants. The remaining three columns show model distributions.}
    \label{fig:pve}
\end{figure}
We compare the probability distributions predicted by all three models against the empirical distribution collected from the population for all of the SLPs and find that the prototype model best matches the empirical distribution. The average of the mean squared errors (MSE) are 1.400 (variance-weighted multi-output $R^2 = 0.609$) for the prototype model, 2.584 (variance-weighted multi-output $R^2=0.284$) for the 1NN exemplar model, and 2.639 (variance-weighted multi-output $R^2=0.299$) for the 2NN exemplar model. In Figure~\ref{fig:pve} we visualize the empirical distribution and the three model distributions for four SLPs.
\subsection{Machine Learning}
The human responses collected in this dataset can be used for supervised training of machine-learning models. The motivation for doing this is three-fold: a) the results can provide evidence for or against human ability to learn in the LO-shot setting; b) a successfully trained model could be used to simulate or predict human behavior which in turn can guide future data collection; and c) it enables a new human-in-the-loop method for training and aligning machine learning systems in extreme low-data regimes without having to explicitly encode strong inductive biases into the models.

We establish a proof-of-concept benchmark by fitting a Random Forest classification model to our dataset. The input features to the model consist of a flattened vector describing a single trial: two coordinates corresponding to the location of the two labeled dinosaurs along the one-dimensional manifold, the associated six soft-label values corresponding to the SLP, and the coordinate corresponding to the location of the third (target) dinosaur. The target output consists of the classification the participant made for that third dinosaur during the trial. We use the {\tt RandomForestClassifier} implementation from the {\tt scikit-learn} Python package~\cite{sklearn_api} with {\tt n\_estimators=20} and {\tt max\_features=None}. 

\begin{figure}[t]
    \centering
    \includegraphics[width=0.99\columnwidth]{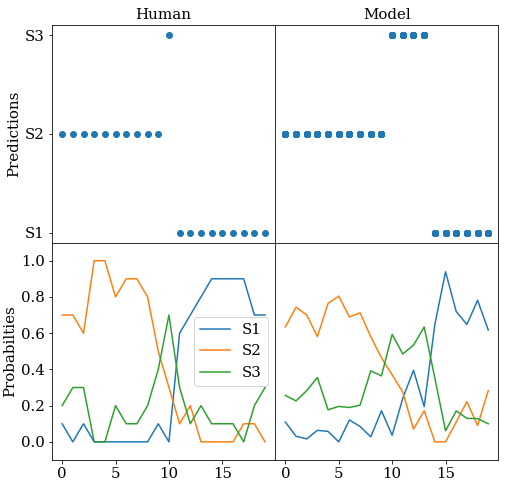}
    \caption{Comparing human behavior with model estimates for each target dinosaur in SLP-13. The model was a random forest classifier trained on human responses from the remaining 13 SLPs. The top row shows the human (majority vote) and model predictions for class membership. The bottom row shows the corresponding probability distributions. }
    \label{fig:ML}
\end{figure}

To avoid over-fitting, we perform leave-one-out cross-validation (LOO-CV) over SLPs which ensures that no participants, nor any SLPs, overlap between training and validation. The population-level classification accuracy averaged over the 14 folds was 88.9\%. The average MSE between the model-estimated probability distribution and the empirical distribution was 0.086 (variance-weighted multi-output $R^2=0.443$). We note that $R^2$ is lower than for the prototype model due to one fold (SLP-4) where the mean values are highly predictive for all classes leading to a large negative $R^2$ value for the random forest model (excluding this fold, average $R^2=0.627$).  These results suggest that machine learning models can indeed successfully predict human behavior in LO-shot settings which reinforces the conclusion that humans act systematically in this low-data regime. We provide an example of human and model predictions in Figure~\ref{fig:ML}.

Promising next steps include expanding the input feature set (e.g. using the stick-figure features or even raw pixels), training other model types (e.g. neural networks), and using trained models to simulate a large dataset over a wide range of the unexplored feature space to guide further data collection.

\section{Discussion}
In this paper, we introduced a novel paradigm for investigating LO-shot learning and soft-label classification in humans. Experimental results from this paradigm 
show that people can learn categories for which they have not seen any exemplars. That is, people's capacity for learning in low-data regimes approaches the theoretical limits of sample efficiency. But which mechanisms enable such efficient learning?

Systematic response patterns (e.g., high between-subject agreement on trials with markedly different superficial features) indicate that participants accurately infer and represent the feature space underlying our generative model. Our modeling suggests that people then form prototypes in this space, and base their final classification judgments on these prototypes, while constraining their inferences through machine-learnable inductive biases.

Though the experiment presented in this paper intentionally used visual stimuli that participants would have little prior knowledge about, our paradigm can conceptually be used to investigate LO-shot learning in other modalities (e.g., classifying auditory stimuli) or domains (e.g., causal learning). Similarly, the category structure underlying our experiment was chosen to be interpretable, but our paradigm enables investigating behavior at higher dimensions and categorical complexity. 

How common are LO-shot learning problems in everyday life? The novel exemplar case analyzed in this paper is likely to occur more frequently in developmental settings: a child might learn that a unicorn is mostly like a horse but also a bit like a rhinoceros, for instance, learning about a novel category without seeing any unicorns in the wild. Adults frequently perform similar inferences as well. If you have never heard of Queen's music before, for instance, your friend might tell you: ``it's a lot like Led Zeppelin and a bit like ABBA too.'' The computational principles outlined in this paper could explain how you can guess what Bohemian Rhapsody might be like without even hearing it. 

More generally, the development of finer-grained conceptual structure may leverage the higher information density provided by soft labels, accessing the full potential of hierarchical inference, especially in domains with fuzzier category boundaries. Future work could investigate whether the models developed in this paper can be generalized to inferences in these domains as well. 

\vspace{2mm}

\noindent {\bf Acknowledgments.} This work was supported by a grant from the John Templeton Foundation.
\bibliographystyle{apacite}

\setlength{\bibleftmargin}{.125in}
\setlength{\bibindent}{-\bibleftmargin}

\bibliography{main}

\end{document}